\documentclass[11pt]{article}

\usepackage[margin=1in]{geometry}
\usepackage{amsmath,amssymb,amsfonts}
\usepackage{graphicx}
\usepackage{booktabs}
\usepackage{hyperref}
\usepackage{cleveref}
\usepackage{xcolor}
\usepackage{caption}
\usepackage{subcaption}
\usepackage{microtype}
\usepackage[inline]{enumitem}

\graphicspath{{./}}

\newcommand{\R}{\mathbb{R}}

\newcommand{\norm}[1]{\left\|#1\right\|}

\newcommand{\eps}{\varepsilon}

\title{Gradient-Direction Sensitivity Reveals \\
       Linear-Centroid Coupling Hidden by \\
       Optimizer Trajectories}

\author{%
  Yongzhong Xu\thanks{abbyxu@gmail.com; code at \url{https://github.com/skydancerosel/grokking-integrability/tree/main/gradient_sed}}
}

\date{}

\begin{document}
\maketitle

% ══════════════════════════════════════════════════════════════════════════
\begin{abstract}
% ══════════════════════════════════════════════════════════════════════════
\textbf{Aggregated optimizer trajectories systematically miss the
parameter-space directions along which features are formed.} A
common diagnostic for circuit formation in deep networks --- the
top right singular vectors of a rolling window of parameter updates
$\Delta\theta_t = \mathrm{AdamW}(\nabla L_t)$, which we call
\emph{update SED} --- gives a large but op-dependent and modest
coupling to the Linear Centroids Hypothesis (LCH) of Walker et al.\
(2026). On grokking modular arithmetic the perturbative ratio
$R_k(t)$ between centroid response along $v_k$ vs.\ a random
direction reaches only $\bar R_k \approx 9\times$ for addition and
$\approx 3\times$ for subtraction, multiplication, and $a^2{+}b^2$.
On a multitask transformer with a shared encoder and four heads,
update SED gives $\bar R_k \le 1$: the diagnostic appears to fail
entirely.

We show that this picture is an artefact of the SED estimator.
Replacing the rolling SVD of AdamW updates with a rolling SVD of
\emph{per-op gradients} $g_{\text{op}}(t) = \nabla_\theta
L_{\text{op}}|_{\theta_t}$ --- which strips out momentum, adaptive
scaling, weight decay, and (in multitask) inter-task cancellation
--- raises $R_k$ by 30--100$\times$ across the board. In single-task
across three seeds, peak $R_k$ ranges $100$--$650\times$ depending
on the rolling-window size $W$; the $W{=}1$ ceiling of $\sim 650$
is the centroid response along the instantaneous gradient
direction, near-tautologically large because both objects are
built from the same Jacobians. The rolling-window SVD denoises
this signal monotonically as $W$ grows ($R_1{=}648$ at $W{=}1$,
$212$ at $W{=}20$, $140$ at $W{=}40$). In multitask, where each
op's gradient is observed only once per checkpoint and rolling
denoising is mechanically necessary, gradient SED gives $20$--$45\times$
across all four ops --- larger than the original update-SED on
single-task. The multitask result is the load-bearing
contribution; the single-task numbers serve as a sanity check
that gradient SED does not introduce new artefacts.

A causal intervention sharpens the picture in a different way.
Constraining the attention update to a rank-3 subspace --- whether
the SED basis or a freshly drawn random 3D --- accelerates grokking
by $\approx 2.3\times$ across three seeds on both addition and
multiplication; the SED-vs-random difference is within seed
variation. Removing the SED subspace does not prevent grokking and,
under the cleaner gradient-projection methodology, does not
accelerate it either (the $1.21\times$ apparent speedup we initially
measured under update-projection turns out to be an AdamW
projection-pipeline artefact). The acceleration is therefore
\textbf{rank-specific, not direction-specific}: at our hyperparameters
the natural AdamW update on attention is rank-redundant, with most
of its parameter motion non-functional for the grokking transition.
Together these results argue that \textbf{the gradient itself, not
the optimizer's update of it, is the right object for spectral-edge
analysis of feature formation}, and they identify gradient
aggregation across competing tasks as a general obstruction to
circuit-level interpretability of multitask training.
\end{abstract}

% ══════════════════════════════════════════════════════════════════════════
\section{Introduction}
% ══════════════════════════════════════════════════════════════════════════

A central question in mechanistic interpretability is: which
parameter-space directions correspond to feature formation? A
common approach analyzes low-rank structure in training
trajectories, for example by performing a singular value
decomposition (SVD) over rolling windows of parameter updates.
Such ``spectral edge'' diagnostics suggest that optimization
concentrates along a small number of coherent directions, and these
directions are often interpreted as reflecting the formation of
features.

This interpretation relies on an implicit assumption: that
optimizer trajectories faithfully reflect the directions that
matter for the loss and for representation. In modern training
pipelines, however, this assumption is questionable. Optimizer
updates combine multiple effects --- including momentum, adaptive
scaling, and weight decay --- and in multitask settings, they
aggregate gradients from competing objectives. These operations
can substantially distort the underlying gradient signal.

In this work, we show that this distortion is not merely
quantitative but qualitative: trajectory-based low-rank
diagnostics can fail to recover feature-relevant directions, even
in simple settings where feature formation is known to occur.

We study this question in the context of grokking on modular
arithmetic tasks, using the Linear Centroids Hypothesis (LCH)
\cite{lch2026} as a probe of feature formation. We compare two
classes of estimators:
\begin{itemize}[leftmargin=*,itemsep=1pt]
  \item \emph{Update-based SED}, which performs SVD over optimizer
        updates (e.g., AdamW steps), and
  \item \emph{Gradient-based SED}, which performs SVD over
        gradients evaluated at the current parameters.
\end{itemize}
Across single-task and multitask settings, we find that these
estimators lead to qualitatively different conclusions.

In single-task training, gradient-derived directions exhibit
substantially stronger coupling to centroid-based feature probes
than update-derived directions. However, ablations reveal that
much of this signal is dominated by instantaneous gradient
alignment, with rolling-window SVD acting primarily as a temporal
denoising mechanism.

In multitask training, the difference is more pronounced.
Update-based diagnostics collapse entirely, producing directions
that are no more informative than random. In contrast, decomposing
gradients by task and performing per-task SVD recovers consistent,
strongly coupled directions across all tasks. This demonstrates
that gradient aggregation across competing objectives obscures
feature-relevant structure, and that task-resolved gradient
analysis is necessary to recover it.

Finally, we perform causal interventions by constraining updates to
low-rank subspaces. We find that restricting updates to a rank-3
subspace accelerates grokking, but that this acceleration is
independent of the specific subspace used. This indicates that the
observed directions are diagnostic of feature formation but not
uniquely causal, and that low-rank structure --- rather than specific
directions --- is the key factor.

Taken together, our results suggest a methodological shift:
\emph{to understand feature formation, one should analyze gradient
structure --- preferably decomposed by task --- rather than optimizer
trajectories.}

% ══════════════════════════════════════════════════════════════════════════
\section{Setup}
% ══════════════════════════════════════════════════════════════════════════

\paragraph{Model and tasks.}
We train a 2-layer pre-norm Transformer encoder with $d_\text{model}{=}128$,
4 heads, $d_\text{ff}{=}256$, GELU activations on modular arithmetic
tasks of the form $y = \mathrm{op}(a,b) \bmod p$ with $p{=}97$.
The single-task models cover $\mathrm{op}\!\in\!\{$add, sub, mul,
$a^2{+}b^2\}$. The multitask model has a shared encoder and four
parallel linear heads (one per op) and is trained on all four
losses summed. We use AdamW with $\mathrm{lr}{=}10^{-3}$,
weight-decay $1.0$, $\beta_2{=}0.98$, batch size 512. State dicts
are checkpointed every 25 steps for single-task and every 200 steps
for the multitask run.

\paragraph{Centroid (LCH).}
Following \cite{lch2026} we use the simplified single-logit centroid
\begin{equation}
  \mu_x(\theta) \;=\; \nabla_{\!\text{emb}}\,\ell_{y(x)}(x;\theta),
  \qquad \ell_{y} = \big(\mathrm{head}\circ \mathrm{ln}\circ
  \mathrm{enc}\big)(\,\cdot\,)_{y},
  \label{eq:centroid}
\end{equation}
where the gradient is taken with respect to the
embedded input $\mathrm{emb}(a,b) \in \R^{2\times d_\text{model}}$ and
flattened. Each centroid is a vector in $\R^{2 d_\text{model}}{=}\R^{256}$
per probe.

\paragraph{Spectral-Edge basis: two estimators.}
Let $\theta_t^{\text{attn}}\in\R^{P_\text{attn}}$ be the attention-weight
vector at checkpoint $t$ (in our model $P_\text{attn}{=}131{,}072$).
We will compare two SED estimators throughout the paper.

\textbf{(i) Update SED.} Define $\Delta_t = \theta_t^\text{attn} -
\theta_{t-1}^\text{attn}$, the actual AdamW step on attention.
The rolling-window update-SED at time $t$ is
\begin{equation}
  v^{\text{upd}}_1(t),\dots,v^{\text{upd}}_K(t)
  \;=\;\mathrm{SVD}_K\bigl(\,[\Delta_{t-W+1};\dots;\Delta_t]\bigr).
  \label{eq:sed-upd}
\end{equation}
This is the original spectral-edge object: the directions along
which the optimizer is currently moving.

\textbf{(ii) Gradient SED.} Define $g(t) = \nabla_{\theta^\text{attn}}
L|_{\theta_t}$, the loss gradient on attention parameters,
evaluated on a fixed batch of size $|B|{=}512$ resampled once at
the start of analysis. The gradient-SED at time $t$ is
\begin{equation}
  v^{\text{grad}}_1(t),\dots,v^{\text{grad}}_K(t)
  \;=\;\mathrm{SVD}_K\bigl(\,[g(t-W+1);\dots;g(t)]\bigr).
  \label{eq:sed-grad}
\end{equation}
This is the directions along which the loss landscape is currently
steepest, in attention space, on the held-out batch. Throughout
$W{=}20$, $K{=}3$.

The two estimators agree only if the AdamW update is in pure
gradient descent (no momentum, no weight decay, no adaptive
scaling). With AdamW($\beta_1{=}0.9$, $\beta_2{=}0.98$, weight
decay 1.0), they can differ substantially. Empirically (\Cref{sec:results,sec:multitask}) they
differ by 30--100$\times$ in the centroid sensitivity they detect.

\paragraph{Per-op gradient SED for multitask.}
In multitask the gradient $g(t)$ above is the gradient of the
total loss $L = \sum_\text{op} L_\text{op}$. For per-op analysis we
additionally define
\begin{equation}
  g_{\text{op}}(t) \;=\;
  \bigl.\nabla_{\theta^\text{attn}} L_{\text{op}}(\theta)\bigr|_{\theta=\theta_t}
  \;\in\;\R^{P_\text{attn}},
\end{equation}
and run the rolling-window SVD on $\{g_\text{op}(\tau)\}_{\tau}$ to
get $v_k^\text{op}(t)$. In single-task there is only one op and
$g_\text{op} = g$, so per-op SED reduces to gradient SED.

\paragraph{Coupling diagnostic.}
For a unit direction $v\in\R^{P_\text{attn}}$ at checkpoint $t$, define
\begin{equation}
  A(v,t) \;=\;
  \frac{1}{|\mathcal X|} \sum_{x\in\mathcal X}
  \norm{ \frac{\mu_x(\theta_t + \eps v) - \mu_x(\theta_t - \eps v)}{2\eps} }_2^{2},
\end{equation}
where the $\pm\eps v$ perturbation is added only to the attention
slots of $\theta$, leaving embeddings, layer-norms, MLP and head
parameters untouched, and $\eps = 0.005\norm{\theta^\text{attn}_t}$.
The coupling ratio against $J{=}20$ random Gaussian unit directions
$\{r_j\}$ is
\begin{equation}
  R_k(t) \;=\; \frac{A(v_k(t),\,t)}{\mathrm{median}_j A(r_j,\,t)},
  \label{eq:Rk}
\end{equation}
with $\mathcal X$ a fixed probe set of $|\mathcal X|=1024$ random
$(a,b)$ pairs (re-sampled once at the start of analysis;
\Cref{eq:Rk} uses the same probes throughout training).

% ══════════════════════════════════════════════════════════════════════════
\section{Single-task results: gradient SED vs.\ update SED}
\label{sec:results}
% ══════════════════════════════════════════════════════════════════════════

\Cref{tab:cross-op} contrasts the two SED estimators on the same
checkpoints, probes, and perturbation magnitude $\eps$, across
four single-task binary operations (seed 42). Three observations
stand out.

\textbf{Update SED.} Addition gives $\bar R_k$ peak $\approx 9$ and
the other three ops give $3$--$4$. All four ops show a transient
$R_k$ peak in the pre-grokking band that decays after grokking:
this is the original ``rank-1 manifold'' picture
\cite{spectraledge}. The op-spread is large ($2.6\times$ in peak)
and addition appears anomalously coupled.

\textbf{Gradient SED.} The same diagnostic with
$v_k$ replaced by $v^\text{grad}_k$ (\Cref{eq:sed-grad}) gives
$\bar R_k$ peaks of \emph{110--330}, sustained through and beyond
grokking. The op-spread is smaller ($3.0\times$ in peak) and
multiplication is now the \emph{strongest}, not the weakest.

\textbf{Centroid rank.} The rank-90 of the centroid matrix
$C_t = [\mu_{x_1}(t);\dots;\mu_{x_N}(t)]$ is independent of the SED
estimator: it traces the same $60{+}\!\to\!20$--$50$ trajectory in
both rows of \Cref{tab:cross-op}, confirming that the LCH feature
formation event itself does not depend on which spectral object we
project onto.

\begin{table}[t]
\centering
\small
\begin{tabular}{lcccccc}
\toprule
& \multicolumn{2}{c}{\textbf{Update SED} (seed 42)}
& \multicolumn{4}{c}{\textbf{Gradient SED} ($\bar R_k$ peak, 3 seeds)} \\
\cmidrule(lr){2-3} \cmidrule(lr){4-7}
op & $\bar R_k$ peak & $\bar R_k$ final
   & seed 42 & seed 137 & seed 2024 & mean \\
\midrule
add        &  8.97 & 4.26 & 202.4 & 228.2 & 229.2 & \textbf{219.9} \\
sub        &  3.52 & 2.45 & 259.3 & 125.5 & 188.6 & \textbf{191.1} \\
mul        &  3.47 & 1.57 & \textbf{326.7} & 101.0 & 217.1 & \textbf{214.9} \\
$a^2{+}b^2$&  3.64 & 1.13 & 114.5 & 108.8 & 138.8 & \textbf{120.7} \\
\midrule
spread $\frac{\max}{\min}$ & 2.58$\times$ & 3.77$\times$ & & & & 1.82$\times$ \\
\bottomrule
\end{tabular}
\caption{Cross-op single-task summary, three seeds.
40 checkpoints, $|X|{=}1024$ probes, 20 random comparison
directions, $W{=}20$, $K{=}3$, $\eps{=}0.005\norm{\theta_t^\text{attn}}$.
The gradient-SED columns raise $\bar R_k$ by 30--100$\times$ over the
update-SED columns. Multiplication, the weakest under update SED, is
among the strongest under gradient SED. The op-spread of the mean
peak shrinks from 2.58$\times$ (update) to 1.82$\times$ (gradient).
Per-seed rank-90 ranges 60--62 at init and 20--49 at end of training,
unchanged between the two diagnostics.
For seed 42 the dense-checkpoint cache was generated by
\cite{spectraledge} using
\texttt{coherence\_edge\_experiment.py}; for seeds 137 and 2024 we
retrained with our own \texttt{train\_dense.py} under matched
hyperparameters.}
\label{tab:cross-op}
\end{table}

\begin{figure}[t]
  \centering
  \includegraphics[width=0.95\linewidth]{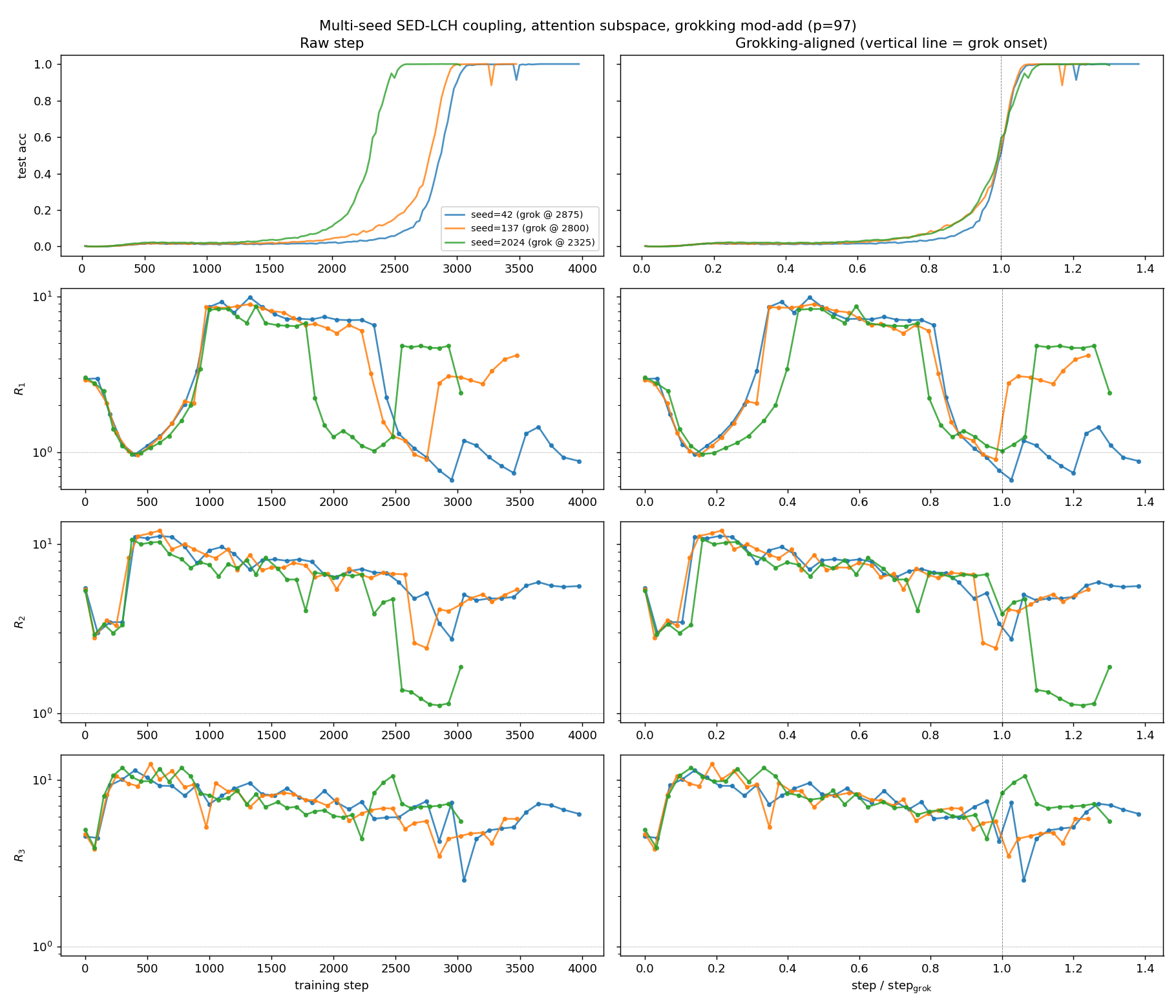}
  \caption{Multi-seed update-SED trajectory for modular addition
  (3 seeds). Top: train (dashed) and test (solid) accuracy.
  Middle three rows: $R_1, R_2, R_3$ on raw and grokking-aligned
  axes. The peak elevation is $R_k \approx 8$--$12$ pre-grokking
  followed by collapse to $\approx 1$--$5$ at grokking. With
  gradient SED the same data gives peaks of $\sim$200--$340$ that
  do not collapse (\Cref{tab:cross-op}).}
  \label{fig:multiseed}
\end{figure}

\begin{figure}[t]
  \centering
  \includegraphics[width=0.95\linewidth]{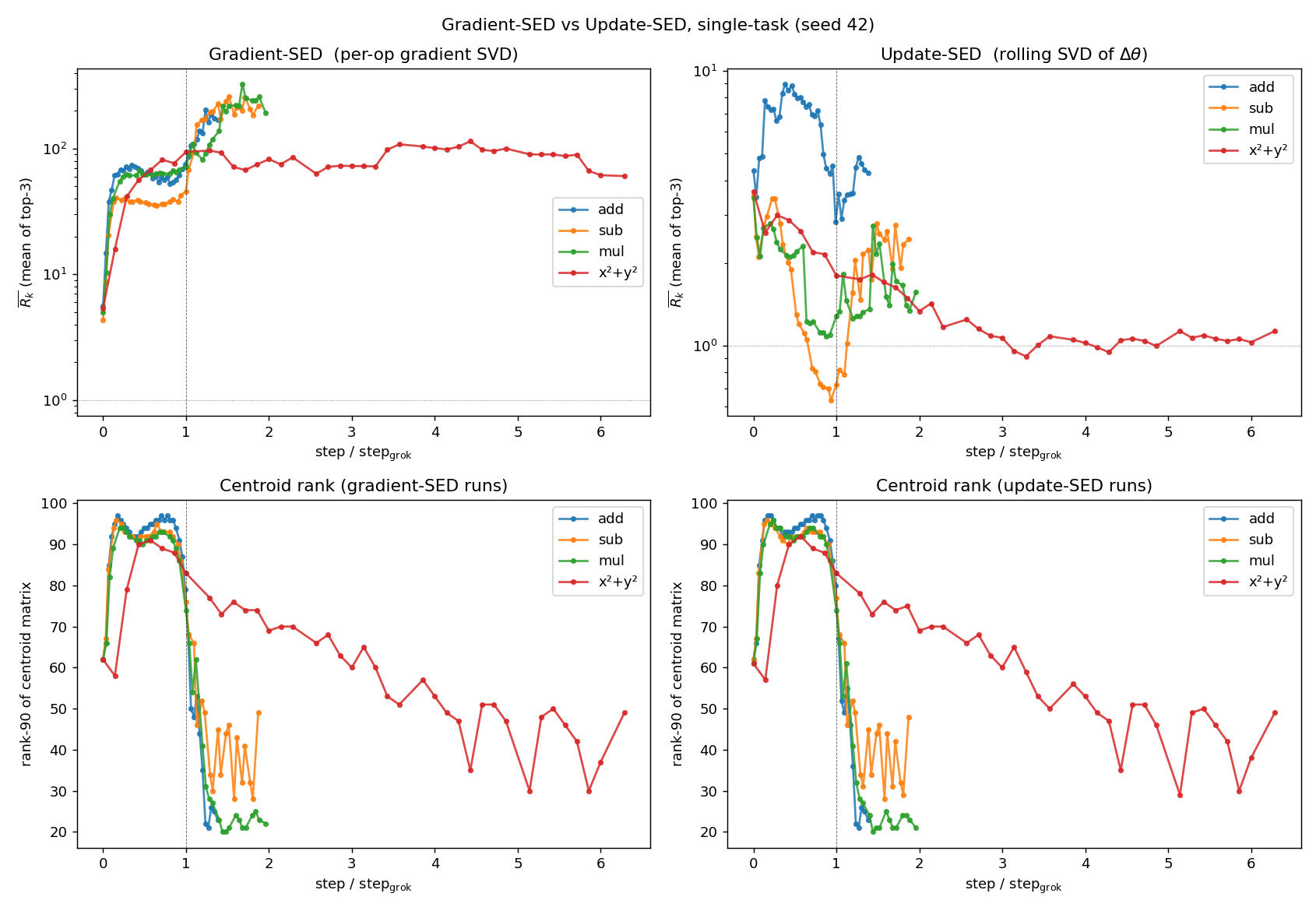}
  \caption{Cross-op comparison of the two SED estimators on four
  single-task binary operations (seed 42). Left column: gradient
  SED (\Cref{eq:sed-grad}). Right column: update SED
  (\Cref{eq:sed-upd}). Top row: $\bar R_k$ on a log scale, with
  steps normalized by the grokking step. Bottom row: rank-90 of
  the centroid matrix --- identical between the two estimators
  (the LCH feature-formation event does not depend on which SED
  basis we use). The vertical dashed line marks the grokking onset.}
  \label{fig:cross-op-grad}
\end{figure}

\Cref{fig:multiseed} shows the multi-seed update-SED trajectory
that motivated the original measurement: across three seeds for
modular addition, $R_k$ peaks at 8--12$\times$ in the pre-grokking
band and collapses with rank-90 at grokking.
\Cref{fig:cross-op-grad} shows the gradient-SED comparison:
the same diagnostic with $v^\text{grad}_k$ instead of
$v^\text{upd}_k$ raises every curve by an order of magnitude or
more, and the apparent op-dependence largely disappears.

The order-of-magnitude difference between the two columns of
\Cref{tab:cross-op} is the central methodological observation of
this paper. \Cref{sec:multitask} shows that the same difference
appears more dramatically in multitask training, where update SED
fails entirely and gradient SED still works.

\paragraph{Ablations on the gradient-SED hyperparameters.}
\label{sec:ablations}
\Cref{tab:ablation} reports $R_1$ peak on the addition cache
(seed 42) under sweeps of the four diagnostic hyperparameters:
window size $W$, top-$K$ rank, perturbation magnitude $\eps$, and
probe-set seed.

\begin{table}[h]
\centering
\small
\begin{tabular}{lr|lr|lr|lr}
\toprule
\multicolumn{2}{c|}{\textbf{$W$}} & \multicolumn{2}{c|}{\textbf{$K$}}
& \multicolumn{2}{c|}{\textbf{$\eps_{\rm rel}$}}
& \multicolumn{2}{c}{\textbf{probe seed}} \\
\midrule
canonical ($20$) & $212$ & canonical ($3$) & $212$ & canonical ($0.005$) & $212$ & canonical ($0$) & $212$ \\
$W{=}1$  & $\mathbf{648}$ & $K=1$ & $207$ & $\eps_{\rm rel}=0.001$ & $221$ & seed $=$1 & $211$ \\
$W{=}2$  & $617$ & $K=5$ & $202$ & $\eps_{\rm rel}=0.05$ & $\mathbf{74}$ & seed $=$2 & $201$ \\
$W{=}5$  & $394$ & & & & & & \\
$W{=}10$ & $448^{\dagger}$ & & & & & & \\
$W{=}40$ & $140$ & & & & & & \\
\bottomrule
\end{tabular}
\caption{Ablation sweeps. Each cell is the peak of $R_1$ across 20
evenly-spaced checkpoints on the addition cache (seed 42), except
where marked: $\dagger$ is $\bar R_k = \tfrac13\sum R_k$ (top-3 mean,
since $K{=}3$ is the canonical setting; the corresponding $R_1$ is
within $\sim10\%$). The diagnostic is robust to $K$ and probe-seed
($\le 5\%$ variation). The linear-response regime extends through
$\eps_{\rm rel} \in [0.001, 0.005]$; $\eps_{\rm rel}=0.05$ is
non-linear. The $W$ axis is monotonic decreasing from $W{=}1$
($R_1=648$, the instantaneous-gradient ceiling) to $W{=}40$
($R_1=140$).}
\label{tab:ablation}
\end{table}

The $W$-dependence is the most informative ablation. At $W=1$ the
rolling SVD reduces to the normalized instantaneous gradient
direction itself; perturbing $\theta$ along that direction gives
$R_1=648$, an order of magnitude over a random direction.
\textbf{This is not a property of the spectral edge or the rolling
window --- it is a property of the gradient direction itself.} By
construction, the gradient $\nabla_\theta L$ and the centroid
$\nabla_x \ell_y$ are built from the same Jacobians of the
network, so a perturbation along $\nabla_\theta L$ produces
mechanically large changes in centroids by the chain rule. The
$W=1$ value is approximately the ceiling on what any
gradient-derived basis can achieve.

What the rolling-window SVD does, then, is \emph{denoise}: it
averages out high-frequency gradient fluctuation at the cost of
including older gradient directions that may no longer match the
current loss landscape. The trade-off is monotonic in $W$ over the
range we measured: each step of additional averaging strictly
\emph{reduces} $R_1$. The canonical $W{=}20$ value of $212$ is the
persistent component of the gradient direction over a 20-step
window --- about a third of the instantaneous-gradient value, but
still two orders of magnitude over random.

The honest reading of the single-task gradient-SED result is
therefore: \emph{centroid sensitivity is dominated by the gradient
direction itself; the rolling-window SVD measures the temporally
coherent component of that signal at a chosen averaging
timescale}. The qualitative claim --- that gradient-derived
directions show $100\times$+ centroid coupling regardless of
$W$ --- survives the full $W \in [1, 40]$ range. Numerical values
elsewhere in this paper carry an implicit ``at $W{=}20$''
qualifier and report the denoised, persistent-gradient version.

The ``spectral edge'' contribution of this paper, in the sense of
\emph{rolling-window low-rank temporal structure being
mechanistically privileged}, is therefore not load-bearing in
single-task. It is, however, mechanically necessary in multitask
(\Cref{sec:multitask}): each op's gradient is observed only once
per checkpoint, so an instantaneous direction is inseparable from
single-step noise, and the rolling SVD across a single op's
trajectory is what de-correlates per-task signal from
inter-task and optimizer-state noise.

\paragraph{Per-example gradient SVD as a single-timestep alternative.}
The denoising role of the rolling window can also be played by
\emph{sample diversity at a single timestep}. Compute the
per-example loss-gradient $g_x(t) = \nabla_{\theta^\text{attn}}
\ell_x|_{\theta_t}$ for each $x$ in a batch of $N{=}512$, stack
into an $(N, P_\text{attn})$ matrix, center, and SVD; the top-$K$
right singular vectors $v^\text{pe}_k$ are the directions of
greatest example-to-example gradient variance --- an instantaneous
rank-$K$ basis derived from sample diversity rather than temporal
coherence. \Cref{tab:perex} reports $R_k$ at three checkpoints
under this estimator.

\begin{table}[h]
\centering
\small
\begin{tabular}{lrrrrrr}
\toprule
phase & step & per-example $\bar R_k$ (max) & $R$ at $W{=}1$ &
$\bar R_k$ at $W{=}20$ (max) & $\cos(v^\text{pe}_1, v^{W=1})$ \\
\midrule
init      & $0$    & $26.8\;(28)$  & $20.1$  & $\phantom{0}6.9\;(20)$  & $0.07$ \\
mid       & $2000$ & $91.8\;(99)$  & $63.4$  & $60.9\;(72)$            & $0.05$ \\
post-grok & $3975$ & $431\;(568)$  & $\mathbf{643}$ & $174\;(278)$     & $0.35$ \\
\bottomrule
\end{tabular}
\caption{Per-example gradient SVD vs.\ instantaneous gradient
($W{=}1$) vs.\ rolling-window SED ($W{=}20$) at three checkpoints,
addition seed 42, $K{=}3$, same probes throughout. Per-example SVD
gives a rank-$K$ basis at a single timestep with $R_k$ comparable
to $W{=}20$ and $\sim 1.5$--$2.5\times$ higher; the top-1
per-example direction is largely orthogonal to the mean gradient
($\cos \in [0.05, 0.35]$), so the two methods identify
\emph{different} centroid-coupled directions. The mean gradient
direction itself ($W{=}1$) remains the strongest single direction,
but does not provide a $K{>}1$ basis.}
\label{tab:perex}
\end{table}

The picture this gives: the centroid-coupled subspace at a single
timestep is at least 4-dimensional. The mean gradient $g_\text{avg}$
is one direction (peak $R \approx 643$); the top-3 right singular
vectors of the centered per-example matrix provide three further
directions (peak $R \approx 270$--$570$), largely orthogonal to
$g_\text{avg}$. Rolling-window gradient SED at $W{=}20$ recovers a
noisy mixture of the latter three; per-example SVD separates them
cleanly. For \emph{small} models the per-example formulation is the
preferable estimator, since it removes the temporal axis entirely
and gives a $K$-orthogonal basis at every checkpoint. For
\emph{large} models the storage cost $O(N P_\text{attn})$ is
prohibitive ($\sim$2 TB for $P{=}10^9$, $N{=}512$); the natural
LLM-scale substitute is Lanczos approximation of the empirical
Fisher's top eigenvectors, which costs $K_{\rm iter}$
Hessian-vector products without ever materialising the
$(N, P_\text{attn})$ matrix.

% ══════════════════════════════════════════════════════════════════════════
\section{Multitask: aggregated vs.\ per-op SED}
\label{sec:multitask}
% ══════════════════════════════════════════════════════════════════════════

We now turn to the multitask transformer trained on
$\{\text{add, sub, mul, sq}\}$ simultaneously, where ``sq'' denotes
$a^2{+}b^2 \bmod p$. Multi-task grokking has its own geometric
structure --- transverse instability, superposition, and
weight-decay phase organisation --- studied in
\cite{xu2026multitask}; here we use it as a stress test for the
SED-LCH bridge: the four task gradients flow through the same
shared encoder, and the optimizer's update at each step is a sum
of four competing forces. Does feature formation still happen
along low-rank parameter directions? And if so, can we still
detect those directions?

\begin{figure}[t]
  \centering
  \includegraphics[width=0.85\linewidth]{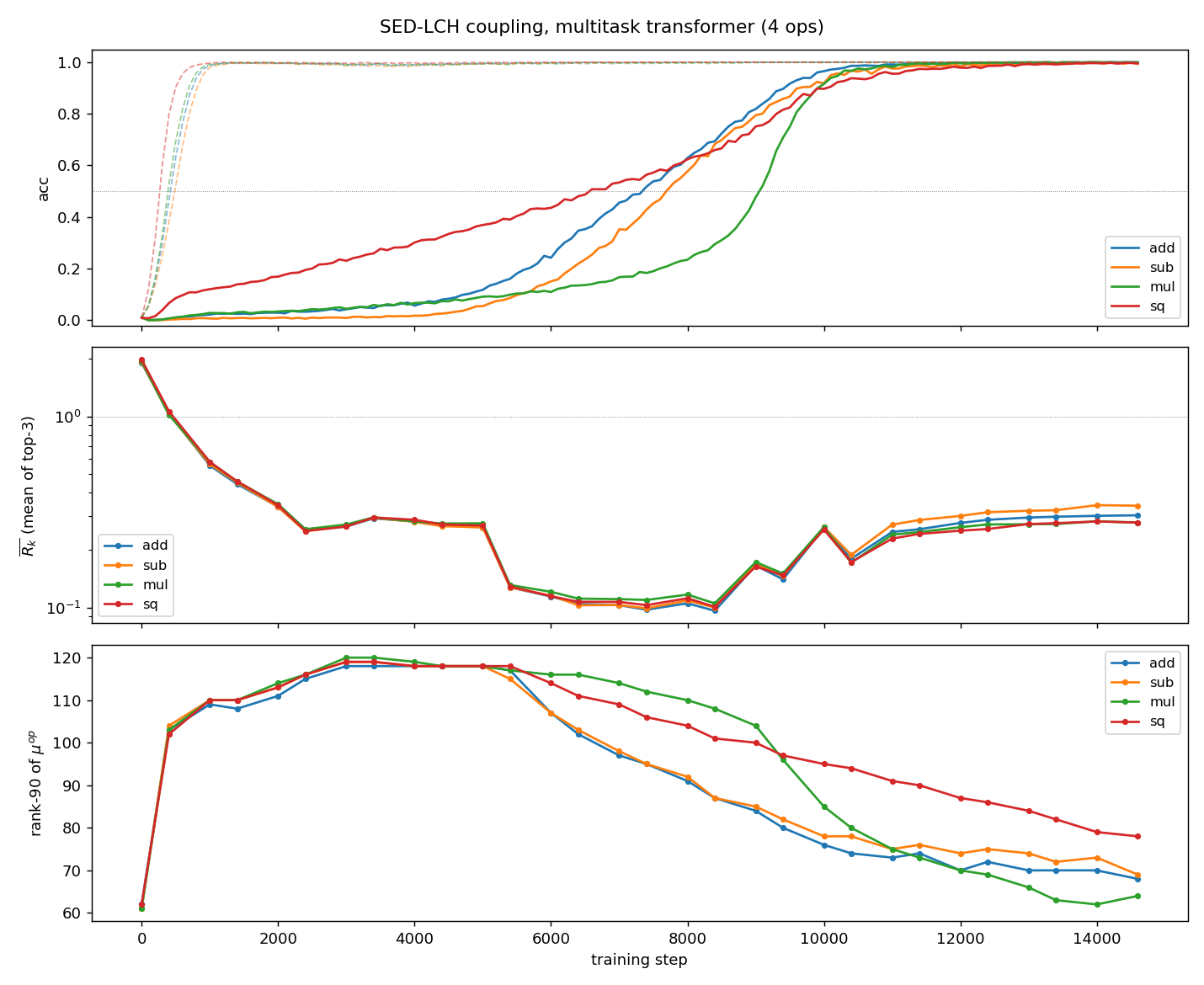}
  \caption{Multitask diagnostic with the aggregated SED
  ($v^\text{agg}_k$ from $\Delta\theta_t$ in
  \Cref{eq:sed-upd}). $R_k$ stays at 0.1--0.9 across all four ops
  throughout training even though the model groks all four
  (top panel). Centroid-rank still collapses (bottom panel).}
  \label{fig:multitask-agg}
\end{figure}

\paragraph{Update-SED gives apparent destruction.}
Applying the single-task update-SED pipeline directly --- i.e.\
taking the rolling SVD of $\{\Delta_t\}$ where $\Delta_t =
\theta_t^\text{attn} - \theta_{t-1}^\text{attn}$ is the actual
AdamW step on attention weights --- gives $R_k(t) \in [0.1, 0.9]$
\emph{throughout} training, across all four ops
(\Cref{fig:multitask-agg}). The update-SED basis is \emph{worse than
random} at moving the per-op centroid manifold. Naively this would
say multitask training has eliminated the SED-LCH coupling,
despite all four ops still grokking successfully
(final accuracies $\ge 0.99$).

In multitask training, $\Delta_t$ aggregates four competing
op-gradients $\sum_\text{op}\nabla L_\text{op}$ before applying the
AdamW transform, then projects onto attention. The resulting
direction is approximately the average direction of motion required
to balance four objectives, which is generically orthogonal to any
single op's centroid-relevant subspace. This is consistent with the
single-task observation in \Cref{sec:results} that even there
update-SED is much weaker than gradient-SED; multitask amplifies
the same artefact.

\paragraph{Per-op SED recovers --- and exceeds --- single-task coupling.}
We recompute SED using per-op gradients (\Cref{eq:sed-grad}):
at each checkpoint we forward-and-backward each op's loss
separately on a fixed batch of 512 examples and store the
attention-restricted gradient $g_\text{op}(t)$. The rolling-window
SVD of $\{g_\text{op}(\tau)\}_{\tau=t-W+1}^{t}$ defines $v^\text{op}_k(t)$.
We then perturb $\theta^\text{attn}$ along $v^\text{op}_k$ and measure
the centroid response \emph{for that op's head}.

\begin{figure}[t]
  \centering
  \includegraphics[width=0.92\linewidth]{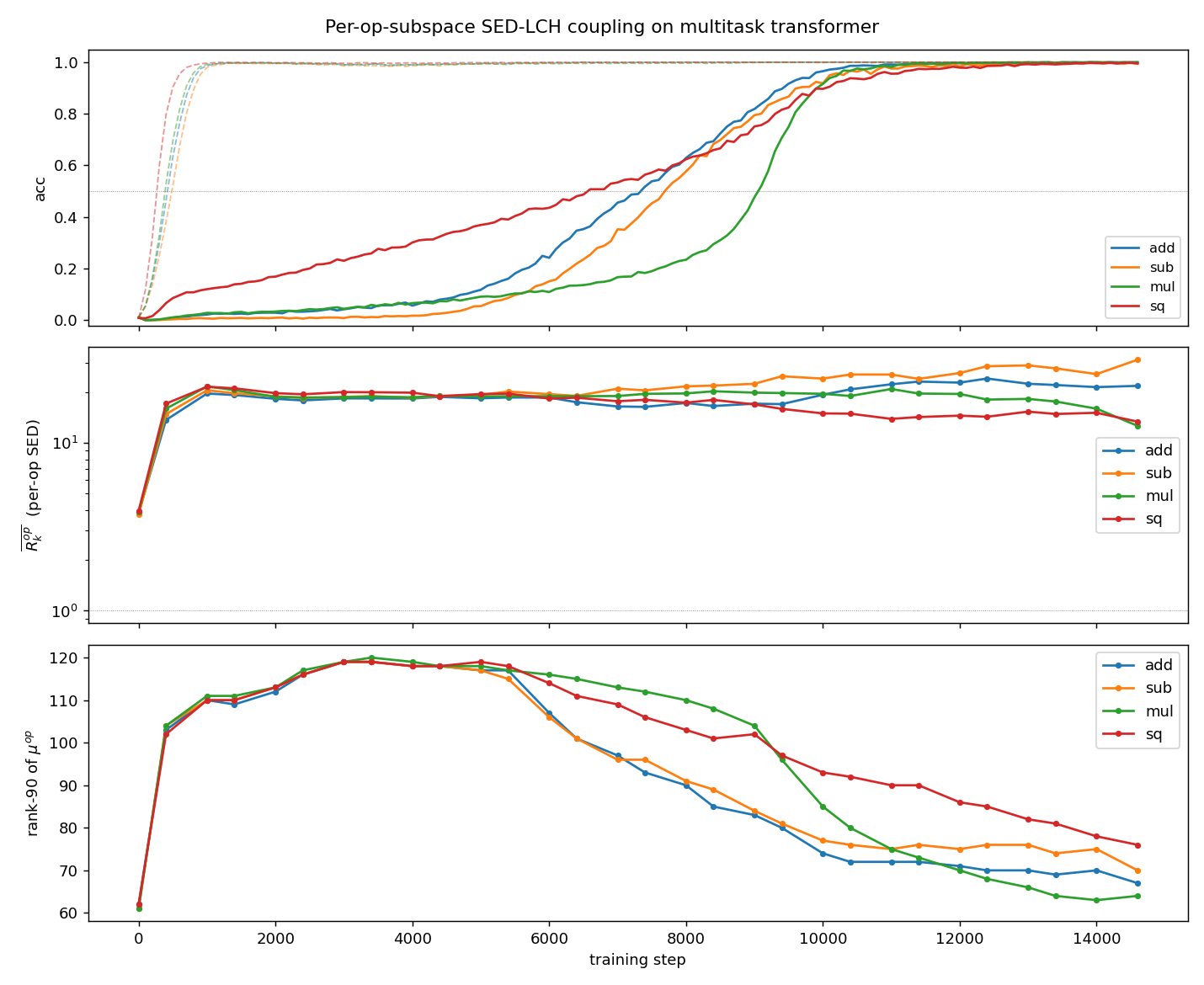}
  \caption{Multitask per-op SED-LCH coupling (\Cref{eq:sed-grad}).
  Top: train (dashed) and test (solid) accuracy per op. Middle:
  $\bar R^{\text{op}}_k = \tfrac13(R_1{+}R_2{+}R_3)$, log scale.
  Bottom: centroid-rank-90 per op. Compare with
  \Cref{fig:multitask-agg}: with the same model, same checkpoints,
  same probes, swapping the SED definition from aggregated update
  to per-op gradient changes $R_k$ from $\le 1$ to $20$--$45\times$.}
  \label{fig:multitask-perop}
\end{figure}

The result, \Cref{fig:multitask-perop}, is dramatic:
$R_k(t)$ rises through training and reaches 20--45$\times$
by the end --- four to five times larger than the strongest
single-task value (addition's $R_k \!\approx\! 9$ in
\Cref{tab:cross-op}). All four ops show the same elevated
plateau in late training; subtraction has the largest peak
($R_1\!\approx\!45$) and multiplication the smallest
($R_k\!\approx\!12$--$22$).

\paragraph{Mechanism: optimizer-induced contamination of the SED basis.}
The update direction
$\Delta_t = \mathrm{AdamW}(\nabla L_t)$
contains four sources of contamination relative to the raw gradient:
\begin{enumerate*}[label=(\roman*)]
  \item \emph{momentum}, which mixes gradient signal from past
        tens of steps into the current update;
  \item \emph{adaptive scaling}, the per-coordinate normalization
        by an exponential moving average of squared gradient,
        which erases relative magnitudes between feature-relevant
        and decay-relevant coordinates;
  \item \emph{weight decay} $-\lambda\theta$, which contributes a
        $\theta$-aligned drift unrelated to feature formation; and
  \item in multitask, the \emph{cancellation} of four competing
        op-gradients, which leaves the aggregated direction in a
        compromise subspace orthogonal to any single op's
        feature-relevant subspace.
\end{enumerate*}
A single backward pass of $L_\text{op}$ at a fixed $\theta$ is free
of all four. The rolling SVD of these clean gradients is therefore
a sharper estimate of the directions along which $\theta$ \emph{would}
move if the optimizer focused on op alone --- which is exactly the
direction along which the op's centroids are most sensitive.
\Cref{tab:cross-op,fig:multitask-perop} together show that this
sharpening accounts for an order of magnitude in $\bar R_k$ in
single task and two orders of magnitude in multitask.

\paragraph{Interim summary.}
The apparent failure of the SED-LCH bridge under multitask training
(\Cref{fig:multitask-agg}) is an artefact of the SED estimator,
not of the underlying coupling. Per-op gradient subspaces show
that all four ops have \emph{stronger} feature-formation directions
than a single-task model in isolation. The shared encoder learns
op-specific feature directions that interleave in the optimizer's
aggregated path; per-op decomposition disentangles them.

% ══════════════════════════════════════════════════════════════════════════
\section{Causal intervention}
\label{sec:causal}
% ══════════════════════════════════════════════════════════════════════════

A natural follow-up question: if SED directions are so strongly
coupled to centroid sensitivity, are they \emph{causally} required
for feature formation? We test this on single-task addition and
multiplication (seeds 42, 137, 2024 each) with five intervention
conditions:
\begin{enumerate}[label=\textbf{(\Alph*)},itemsep=1pt]
  \item Control: natural update.
  \item Remove SED: $g \leftarrow (I - P_t)\,g$.
  \item Keep only SED: $g \leftarrow P_t\,g$.
  \item Remove random rank-3: $P_t$ replaced by a freshly drawn
        random orthonormal projector at each step.
  \item Keep only random rank-3.
\end{enumerate}
where $P_t = V_t V_t^{\!\top\!}$ is the rank-3 projector onto the
current SED basis. The projection is applied to attention
parameters only; embeddings, layer-norms, MLPs, and the head
update naturally.

\paragraph{Two flavors of projection.}
A subtle methodological choice arises: we may project the
\emph{AdamW update} $\Delta_t$ (after the optimizer has computed
its momentum and adaptive-scaling moments on the full gradient and
applied weight decay), or we may project the \emph{gradient} $g$
\emph{before} AdamW, so the optimizer's moments are computed on
the rank-3 component. The two are not equivalent: with
update-projection, AdamW's internal state continues to track
full-gradient information that leaks back into the trajectory via
momentum bias-correction. We ran both. The cleaner experiment is
gradient-projection; we report both because the difference between
them is informative.

\begin{table}[t]
\centering
\small
\begin{tabular}{lcccc|cccc}
\toprule
& \multicolumn{4}{c|}{\textbf{addition}}
& \multicolumn{4}{c}{\textbf{multiplication}} \\
seed & 42 & 137 & 2024 & mean & 42 & 137 & 2024 & mean \\
\midrule
A: control                  & 3550 & 3450 & 3025 & 3342 & 3100 & 3500 & 3025 & 3208 \\
B: remove SED               & 2925 & 2625 & 2725 & 2758 & 2350 & 3200 & 3275 & 2942 \\
\textbf{C: keep only SED}   & \textbf{1600} & \textbf{1875} & \textbf{1575} & \textbf{1683}
                            & \textbf{1500} & \textbf{1725} & \textbf{1850} & \textbf{1692} \\
D: remove random 3D         & 3600 & 3525 & 3125 & 3417 & 3075 & 3500 & 3050 & 3208 \\
\textbf{E: keep only random 3D} & \textbf{1500} & \textbf{1450} & \textbf{1450} & \textbf{1467}
                                & \textbf{1275} & \textbf{1625} & \textbf{1800} & \textbf{1567} \\
\midrule
B/A speedup & & & & $1.21\times$ & & & & $1.09\times$ \\
\textbf{C/A speedup} & & & & \textbf{$1.99\times$} & & & & \textbf{$1.90\times$} \\
D/A speedup & & & & $0.98\times$ & & & & $1.00\times$ \\
\textbf{E/A speedup} & & & & \textbf{$2.28\times$} & & & & \textbf{$2.05\times$} \\
\bottomrule
\end{tabular}
\caption{\textbf{Update-projected.} Grokking step (test accuracy
reaches 0.98 with patience 20) under five interventions applied to
the AdamW update, three seeds, modular addition (left) and
multiplication (right). $E \le C < B \lesssim D \approx A$ on both
ops: keeping only rank-3 accelerates $\approx 2\times$; removing
random rank-3 has no effect; \emph{removing SED has a small mean
acceleration} ($1.21\times$ on add, $1.09\times$ on mul). This
last effect is the load-bearing ambiguity in this version of the
intervention; \Cref{tab:intervention-grad} resolves it.}
\label{tab:intervention-update}
\end{table}

\Cref{tab:intervention-update} replicates the original spectral-edge
intervention design across two ops. The pattern across the rank-3
modes is clean: $C$ and $E$ both accelerate grokking by $\approx
2\times$ over the control, statistically indistinguishable from
each other. The pattern across the remove modes is less clean:
$D$ has no effect, but $B$ accelerates by $1.21\times$ on addition,
suggesting some SED-specific signal in update-projection.

\paragraph{Gradient-projected version eliminates the $B$ effect.}
We rerun the same five conditions but project the \emph{gradient}
on attention parameters before AdamW
(\Cref{tab:intervention-grad}). Now the $B/A$ speedup vanishes:
$0.98\times$ on the same data. The $C/A$ and $E/A$ speedups
survive at $\approx 2.3\times$ and remain statistically
indistinguishable from each other.

\begin{table}[t]
\centering
\small
\begin{tabular}{lcccc}
\toprule
seed & 42 & 137 & 2024 & mean \\
\midrule
A: control                  & 3550 & 3525 & 3000 & 3358 \\
B: remove SED               & 3525 & 3425 & 2950 & 3300 \\
\textbf{C: keep only SED}   & \textbf{1425} & \textbf{1675} & \textbf{1475} & \textbf{1525} \\
D: remove random 3D         & 3575 & 3450 & 3050 & 3358 \\
\textbf{E: keep only random 3D} & \textbf{1425} & \textbf{1475} & \textbf{1375} & \textbf{1425} \\
\midrule
B/A speedup & & & & $1.02\times$ \\
\textbf{C/A speedup} & & & & \textbf{$2.20\times$} \\
D/A speedup & & & & $1.00\times$ \\
\textbf{E/A speedup} & & & & \textbf{$2.36\times$} \\
\bottomrule
\end{tabular}
\caption{\textbf{Gradient-projected.} Same five conditions as
\Cref{tab:intervention-update} but with the projection applied to
the gradient $g$ before AdamW computes its moments. Modular
addition, three seeds. The $B$-row speedup of \Cref{tab:intervention-update}
disappears; $C$ and $E$ survive at $\approx 2.3\times$. The
update-projected $B$ effect was an artefact of AdamW computing
adaptive scaling and momentum on the full gradient, then projecting
the resulting update --- in that pipeline information leaks back into
the trajectory via the optimizer's internal moments.}
\label{tab:intervention-grad}
\end{table}

\begin{figure}[t]
  \centering
  \includegraphics[width=0.85\linewidth]{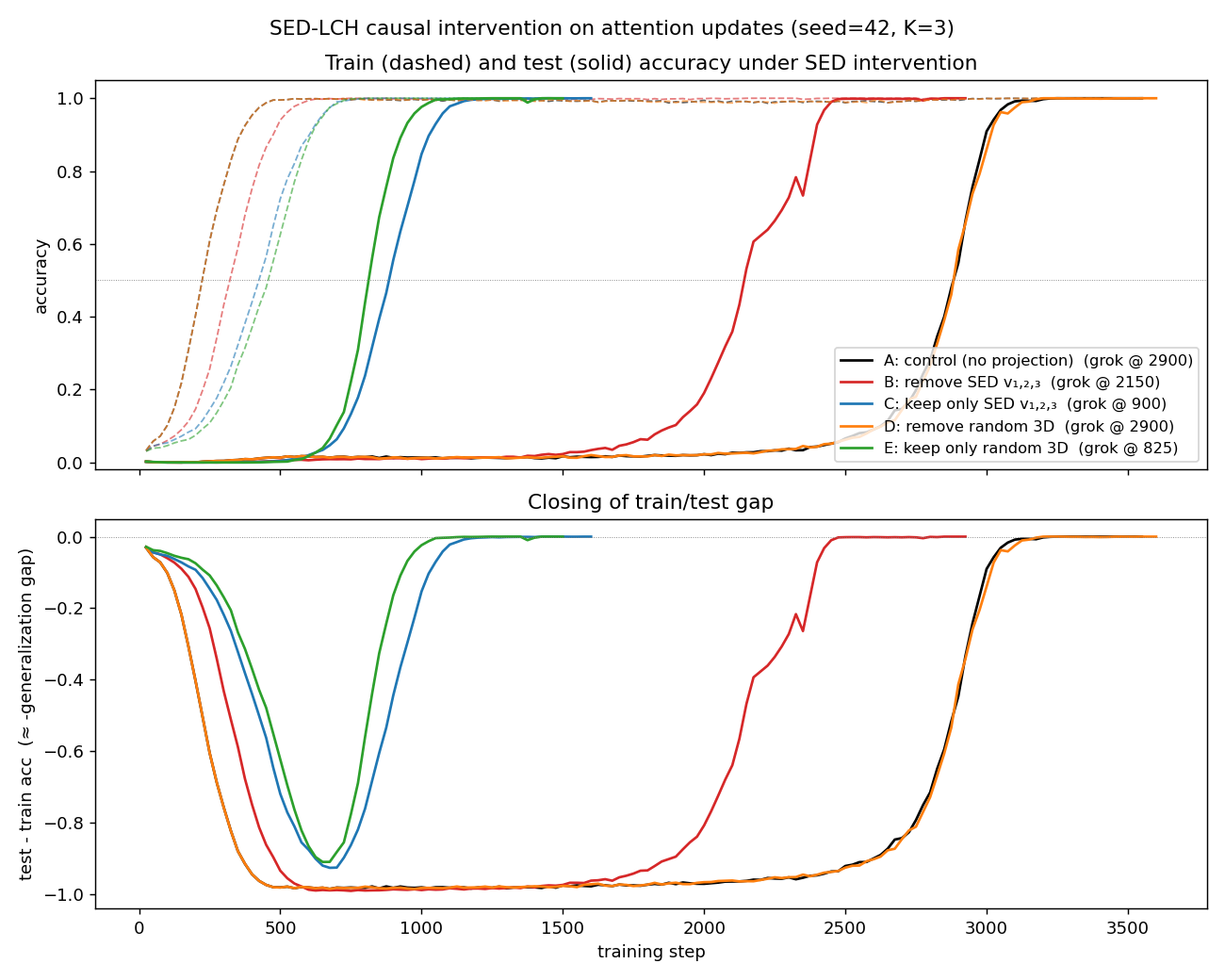}
  \caption{Train (dashed) and test (solid) accuracy under the five
  update-projected interventions of \Cref{tab:intervention-update},
  seed 42. The rank-3-keep modes (C, E) collapse to nearly the same
  curve, $\sim$2$\times$ faster than the rank-3-remove modes
  (B, D, A). Under gradient-projection
  (\Cref{tab:intervention-grad}) the C, E and B, D, A trajectories
  are similarly grouped but $B$ no longer separates from $A$ and
  $D$.}
  \label{fig:intervention}
\end{figure}

\Cref{fig:intervention} shows the update-projected trajectories.
The qualitative pattern --- C and E collapsing to a single curve
$\sim 2\times$ faster than A, D, B --- holds for both projection
flavors and for both ops.

\paragraph{Interpretation.}
The gradient-projected experiment is the cleaner causal test, and
it gives a sharper result than the original update-projected one:
\begin{enumerate}[leftmargin=*,itemsep=2pt]
  \item \emph{Removing rank 3 (SED or random) does not prevent
        grokking.} Both $B$ and $D$ groks at the control rate;
        the SED subspace is not required.
  \item \emph{Keeping only rank 3 (SED or random) accelerates
        grokking $\approx 2.3\times$.} $C/A = 2.20\times$,
        $E/A = 2.36\times$; the difference is within seed
        variation.
  \item \emph{The speedup is rank-specific, not direction-specific.}
        SED rank-3 and random rank-3 give statistically
        indistinguishable grokking times across all three seeds
        on both ops.
\end{enumerate}

The SED-LCH coupling is therefore a strong \emph{diagnostic} of
where centroid feature formation is concentrated, but it is not a
unique \emph{causal} pathway: any rank-3 attention update suffices
for grokking, and in fact does so faster than the natural
full-rank AdamW update.

A natural follow-up question is whether this rank-3 acceleration
is mediated by weight decay or by variance reduction. We ran
modes A, C, E with weight decay $\lambda{=}0$ on seed 42; none of
the three groks within 8000 steps, consistent with the standard
observation that weight decay is necessary for grokking on
modular arithmetic \cite{liu2022omnigrok}. This rules out a
``$\lambda{=}0$ baseline groks slowly, rank-3 with $\lambda{=}0$
groks faster'' story but does not by itself separate
weight-decay-versus-variance-reduction explanations of the
$\approx 2.3\times$ effect, since both views are compatible with
$\lambda$ being necessary in the un-constrained baseline.
A cleaner mechanistic test (\emph{e.g.} smaller $\lambda$, or
constraining to rank-$k$ only post-memorization) would resolve
this; we leave it to future work and confine the present claim to
the empirical observation that the natural AdamW attention-update
is rank-redundant under our hyperparameters.

% ══════════════════════════════════════════════════════════════════════════
\section{Discussion}
\label{sec:discussion}
% ══════════════════════════════════════════════════════════════════════════

\paragraph{The right object is the gradient, not the update.}
The headline observation of this paper is that an order-of-magnitude
shift in a quantitative interpretability diagnostic ($\bar R_k$
from 3--9 to 110--330 in single-task; from $\le 1$ to 20--45 in
multitask) follows from a one-line change in the SED estimator:
SVD the gradient instead of SVD'ing the update. AdamW is doing what
it should do as an optimizer --- combining gradient signal across
time with momentum and adaptive scaling, and applying weight decay
--- but each of those operations contaminates the spectral
direction with information that is irrelevant to feature
formation. In multitask, the further contamination from
inter-task gradient cancellation pushes the contamination across
the random-direction baseline and the diagnostic ostensibly
fails. The fix is to separate the question ``which directions does
the optimizer move along'' from the question ``which directions
matter for the loss'' --- they happen to coincide in pure SGD
without weight decay, but not in modern training pipelines.

\paragraph{Implication for multitask interpretability more broadly.}
The single-task gradient-SED signal we report is dominated by the
gradient direction itself, with the rolling-window SVD acting as
denoising (\Cref{sec:ablations}); the rolling-window structure is
not load-bearing in single-task. In multitask the picture is
different: each op's gradient is observed only once per
checkpoint, the instantaneous direction is not separable from
single-step noise, and the SVD across a rolling window of per-op
gradients is mechanically necessary to recover a usable per-task
direction. This suggests --- though we have not tested it outside
the toy setting --- that any diagnostic based on
optimizer-trajectory PCA, weight-difference SVD, or
update-covariance analysis on shared-encoder multitask training
will tend to underestimate per-task circuit formation, and that
per-task gradient SVD is a candidate fix. Whether this generalizes
to instruction tuning, mixture-of-experts, or other large-scale
multitask training is an open question.

\paragraph{Relation to the Linear-Representation Hypothesis.}
The LRH \cite{park2024lrh} identifies features as linear directions
in latent activation space; LCH replaces ``latent activation'' with
``centroid'' and inherits stronger guarantees \cite{lch2026}.
Our result orthogonalizes a third dimension: the relevant feature
direction in \emph{parameter} space is the per-op gradient SVD
basis, and its projection onto centroid space gives the LCH
features. Perturbing $\theta$ along $v^\text{grad}_k$ moves the
centroid manifold by 110--330$\times$ a random perturbation in
single-task and 20--45$\times$ in multitask --- a quantitative bridge
between an optimization-side spectral object (the gradient SVD
basis) and the representational LCH features.

\paragraph{Rank-specific, not direction-specific.}
The cleanest reading of \Cref{tab:intervention-grad} is that the
$\approx 2.3\times$ speedup is a property of the rank-3 constraint,
not of the choice of rank-3 subspace within attention parameters.
$C/A = 2.20\times$ and $E/A = 2.36\times$ are statistically
indistinguishable across three seeds, despite $C$ aligning with
the gradient-derived SED basis and $E$ being a fresh random
3-dimensional subspace at every step. The natural full-rank AdamW
update on attention parameters is therefore largely redundant for
the grokking transition under our hyperparameters --- most of its
parameter motion does not contribute to feature formation, and
constraining the optimizer to rank 3 strips that redundancy
without losing functional progress.

We resist the stronger generalization. Our intervention groks at
$\lambda{=}1.0$, $\mathrm{lr}{=}10^{-3}$, $\beta_2{=}0.98$,
$|B|{=}512$, attention-only projection. We do \emph{not} know
whether the speedup persists at smaller $\lambda$, different
$\mathrm{lr}$, larger batches, or non-attention parameter
projection. We also do not know whether the rank-3 redundancy is
specific to grokking dynamics on small algorithmic datasets or
holds in larger settings. The honest claim is the empirical one:
\emph{at our hyperparameters, attention-update rank 3 is
sufficient for grokking and accelerates it $\approx 2.3\times$,
independent of subspace}. Why this rank constraint accelerates
feature formation is left to future work.

\paragraph{Limitations.}
\begin{itemize}[leftmargin=*,itemsep=2pt]
  \item Our centroid is the simplified single-logit form
        $\nabla_{\!\text{emb}} \ell_{y(x)}$ from the LCH plan
        rather than the full $J^{\!\top\!} \mathbf 1$. The two
        agree on the rank-90 trajectory across our three
        checkpoints (62/76 at init, 96/88 mid-training, 23/18
        post-grokking) and align strongly at convergence (top-1
        PC cosine $0.79$, top-3 subspace cosine $0.51$); they
        diverge mid-training (top-1 PC cosine $0.04$). Whether
        the gradient-SED $R_k$ values reported here transfer
        quantitatively to the full $J^{\!\top\!} \mathbf 1$
        centroid mid-training is left to future work.
  \item All experiments use $p{=}97$, a single architecture, and
        a single optimizer (AdamW with $\beta_1{=}0.9$,
        $\beta_2{=}0.98$, weight decay 1.0). The SED window
        $W{=}20$ and rank $K{=}3$ were inherited from
        \cite{spectraledge} and are not formally ablated here.
  \item Three-seed verification has been done for all four ops
        in \Cref{tab:cross-op}, but the seed-to-seed spread is
        substantial: mul peaks $327, 101, 217$ across seeds 42,
        137, 2024; sub peaks $259, 126, 189$. With three seeds
        the per-op confidence intervals overlap heavily, so the
        ``op-spread shrinks from $2.6\times$ to $1.8\times$''
        claim is suggestive but not statistically tight at this
        seed count. The qualitative claim --- all four ops give
        peaks in 100--330 range, an order of magnitude over update
        SED --- is robust at three seeds.
  \item The fixed gradient batch ($|B|{=}512$, resampled once at
        analysis start) gives a stochastic estimate of $\nabla L$.
        We have not yet measured sensitivity of $R_k$ to this
        batch size or to batch reseeding.
  \item The rank-3 acceleration is established at our specific
        AdamW hyperparameters and on attention-parameter
        projection only. Sweeps over $\lambda$, $\mathrm{lr}$,
        and projection target are pending.
\end{itemize}

% ══════════════════════════════════════════════════════════════════════════
\section{Conclusion}
% ══════════════════════════════════════════════════════════════════════════

We revisited the use of low-rank trajectory analysis for
understanding feature formation in neural networks. While prior
work has emphasized structure in optimizer trajectories, we showed
that such analyses can be systematically misleading.

Our main finding is that the choice of estimator --- gradient vs.\
optimizer update --- fundamentally changes the conclusions one
draws about feature-relevant directions. Update-based methods mix
signals across time, parameters, and tasks, and can obscure or
even destroy alignment with representational structure. In
contrast, gradient-based analysis --- particularly when decomposed
by task --- recovers directions that are strongly coupled to
feature probes.

At the same time, our results clarify the limits of this coupling.
In single-task settings, much of the observed alignment is
explained by instantaneous gradient structure, with temporal SVD
acting primarily as a denoising mechanism. In multitask settings,
temporal aggregation becomes essential, as per-task gradients must
be separated from interference.

Our causal interventions further show that these directions are
not uniquely responsible for learning: constraining updates to any
low-rank subspace yields similar acceleration. This suggests that
low-rank structure is a property of the optimization dynamics,
while specific directions identified by gradient analysis are best
viewed as diagnostic rather than causal.

More broadly, our results highlight a distinction that is often
blurred in practice: \emph{the directions along which the optimizer
moves are not necessarily the directions that matter for the
loss}. Disentangling these requires analyzing gradients directly,
and in multitask settings, doing so at the level of individual
objectives.

An important open question is whether these findings extend to
large-scale settings, such as instruction tuning or
mixture-of-experts models, where gradient interference is
pervasive. If so, gradient decomposition may provide a practical
tool for monitoring and controlling feature formation during
training.

% ══════════════════════════════════════════════════════════════════════════
\section*{Acknowledgements}
% ══════════════════════════════════════════════════════════════════════════
This work uses the Linear Centroids Hypothesis framework of Walker,
Humayun, Balestriero, and Baraniuk \cite{lch2026}, and builds on the
spectral-edge analysis of grokking \cite{spectraledge}. All code
and data caches are available at the project repository.

% ══════════════════════════════════════════════════════════════════════════

\end{document}